\documentclass{article} 
\usepackage{iclr2026_conference}

\usepackage{amsmath,amsfonts,bm}

\def\eqref#1{equation~\ref{#1}}

\def\1{\bm{1}}

\DeclareMathAlphabet{\mathsfit}{\encodingdefault}{\sfdefault}{m}{sl}
\SetMathAlphabet{\mathsfit}{bold}{\encodingdefault}{\sfdefault}{bx}{n}

\usepackage[T1]{fontenc}
\usepackage{times}
\usepackage{amsmath, amssymb, amsthm}
\usepackage{mathtools}
\usepackage{bbm}
\usepackage{dsfont}
\usepackage{adjustbox}

\usepackage[dvipsnames]{xcolor}
\definecolor{darkpink}{RGB}{199,21,140}

\usepackage{graphicx}
\usepackage{epstopdf}

\usepackage{booktabs, array}

\usepackage{listings}
\usepackage{fancyvrb}
\fvset{fontsize=\small}

\definecolor{citecolor}{RGB}{0,102,204}
\definecolor{linkcolor}{RGB}{190,105,30}
\definecolor{urlcolor}{RGB}{199,21,133}

\usepackage[colorlinks,linktoc=all]{hyperref}
\usepackage[all]{hypcap}
\hypersetup{citecolor=citecolor}
\hypersetup{linkcolor=linkcolor}
\hypersetup{urlcolor=urlcolor}
\usepackage[nameinlink,capitalise]{cleveref}
\creflabelformat{equation}{#2\textup{#1}#3}  
\crefname{section}{\S}{\S\S}

\lstdefinestyle{mystyle}{
    commentstyle=\color{OliveGreen},
    numberstyle=\tiny\color{black!60},
    stringstyle=\color{BrickRed},
    basicstyle=\ttfamily\scriptsize,
    breakatwhitespace=false,
    breaklines=true,
    captionpos=b,
    keepspaces=true,
    numbers=none,
    numbersep=5pt,
    showspaces=false,
    showstringspaces=false,
    showtabs=false,
    tabsize=2
}
\usepackage{subcaption}
\usepackage{wrapfig}
\usepackage{bbm}
\usepackage{multirow}
\usepackage{makecell}
\usepackage[table]{xcolor}
\usepackage{pifont} 
\usepackage{colortbl}

\usepackage[most]{tcolorbox}
\usepackage{graphicx}
\usepackage{xcolor}
\usepackage{lipsum}

\usepackage{listings}
\tcbuselibrary{listings}

\tcbset{
  highlight token/.style={
    boxrule=0.4pt,
    colback=gray!5,
    colframe=gray!50,
    fontupper=\ttfamily\color{purple},
    arc=1pt,
    boxsep=1pt,
    left=0.5pt,
    right=0.5pt,
    top=0.5pt,
    bottom=0.5pt,
    enhanced,
    box align=base,
    on line,
    height=1.2em,
    valign=center, 
  }
}

\newcommand{\gain}[2]{%
  #1\,{\scriptsize\textcolor{blue!70!black}{#2}}%
}
\newcommand{\loss}[2]{%
  #1\,{\scriptsize\textcolor{red!70!black}{#2}}%
}
\newcommand{\same}[2]{%
  #1\,{\scriptsize\textcolor{gray}{#2}}%
}

\definecolor{promptbg}{RGB}{245,245,245}
\definecolor{promptborder}{RGB}{200,200,200}
\definecolor{prompttitle}{RGB}{40,40,40}
\definecolor{prompttext}{RGB}{20,20,20}

\newtcolorbox[auto counter]{prompt}[2][]{
  listing only,
  listing options={
    basicstyle=\ttfamily\footnotesize\color{prompttext},
    breaklines=true,
    showstringspaces=false,
  },
  colback=promptbg,
  colframe=promptborder,
  coltitle=prompttitle,
  title=>>>~#2,
  fonttitle=\bfseries,
  boxrule=0.4pt,
  arc=2pt,
  left=1em,
  right=1em,
  top=0.7em,
  bottom=0.7em,
  #1,
}

\usepackage{mwe}
\usepackage{siunitx}
\usepackage{url}

\title{Bridging the Missing-Modality Gap:\\
Improving Text-Only Calibration of Vision Language Models}

\author{Mingyeong Kim \hfill Jungwon Choi \hfill Chaeyun Jang \hfill Juho Lee\\
Kim Jaechul Graduate School of AI, KAIST\\
\texttt{\{kmk50411,jungwon.choi,jcy9911,juholee\}@kaist.ac.kr}
}

\iclrfinalcopy 
\begin{document}

\maketitle

\begin{abstract}

Vision-language models (VLMs) are often deployed on text-only inputs, although they are trained with images. We find that removing the vision modality causes large drops in accuracy and severe miscalibration, and the model does not behave like its original language backbone under text-only prompting. This failure is not explained only by missing semantic information. Even when text descriptions preserve key content, confidence becomes unreliable, while adding a visual signal through generated images partially restores accuracy and calibration. We propose the Latent Imagination Module (LIM), a lightweight cross-attention module that predicts imagined latent embeddings from textual input and feeds them into a frozen VLM backbone without pixel-level image synthesis. Across text-only benchmarks, unseen tasks, and missing-image scenarios, LIM improves accuracy and reduces calibration error. These results suggest that latent modality completion is a practical approach for reliable VLM inference under missing-modality.
\end{abstract}

\section{Introduction}
\label{introduction}
\vspace{-0.1em}

Recent large language models (LLMs) have rapidly evolved into vision–language models (VLMs) that jointly process textual and visual inputs. These models have demonstrated strong performance in image-based question answering and multimodal reasoning. They are increasingly adopted in both research and real-world systems due to their ability to integrate visual and linguistic information within a single unified architecture~\citep{liu2023visual, deitke2025molmo, yang2025qwen3}. In particular, recent open-source and commercial models, such as GPT-4o~\citep{hurst2024gpt}, are built around a unified vision–language backbone, where text-only inputs are typically handled by the same multimodal model, rather than by a separate text-only architecture.

Despite their multimodal design, a substantial portion of tasks performed by deployed VLMs remain text-only~\citep{chatterji2025people}. For example, conversational agents, question answering, reasoning, retrieval assistance, and tool-use decision-making are often executed without any visual input. In these settings, not only predictive accuracy but also prediction reliability and calibration play a critical role. Model confidence is frequently used to determine whether to invoke external tools or retrieval systems, whether to abstain from answering, and how to prioritize human review in human-in-the-loop pipelines or safety filtering mechanisms~\citep{jang2025reliable, xu2025alignment}. Consequently, ensuring reliable and well-calibrated confidence predictions in text-only settings is a practically important requirement for VLMs.

However, we find that when deployed in text-only settings, VLMs suffer not only from degraded accuracy but also from severe miscalibration, even compared to their text-only backbones prior to multimodal adaptation. Moreover, this degradation barely improves when we augment text input with additional natural-language descriptions of the missing visual content. In contrast, our experiments show that generating and providing task-relevant images of text-only input can partially restore both accuracy and calibration, highlighting the importance of visual signals in VLM inference even when the original task does not include real images. These results suggest that the performance drop cannot be addressed by purely linguistic augmentation, but instead stems from a mismatch in internal representation induced by multimodal training. This motivates a mechanism that can supply missing visual signals at the representation level without requiring explicit image generation.

To overcome this limitation, we propose the \textbf{Latent Imagination Module (LIM)}. Rather than treating text as a standalone input, our framework guides the model to infer imagined latent embeddings directly from text via cross-attention, without expensive pixel-level generation. These completed latent embeddings are then provided to the unified backbone, enabling text-only inference to induce latent states that more closely resemble those produced by multimodal inputs, thereby stabilizing confidence predictions and improving calibration. While prior work has explored missing-modality learning and VLM calibration 
independently, our work is the first to connect these two threads by identifying 
the missing-modality gap as a source of miscalibration and proposing a latent-space 
solution. We discuss related work in detail in Appendix~\ref{related}.

\vspace{0.5em}
\noindent\textbf{Contributions.} The main contributions of this work are summarized as follows:
\begin{itemize}
    \item We identify the \textit{Missing-Modality Gap} as a primary cause of miscalibration in VLMs under text-only settings, framing it as a cognitive rupture rather than a simple data deficit.
    \item We propose a novel \textbf{Latent Imagination Module} that uses cross-attention to infer task-relevant latent embeddings directly from text, effectively bridging the modality gap without expensive pixel-level reconstruction.
    \item We demonstrate that our end-to-end latent imputation strategy significantly improves calibration and reliability across both \emph{inherently missing-modality settings}, where a required modality is absent, and \emph{modality-mismatched regimes}, where text-only tasks are deployed on multimodal backbones. These results suggest that \emph{latent imagination} acts as a robust regularizer by aligning text-only inference with the multimodal training distribution.
\end{itemize}

\section{Empirical Findings}
\label{findings}

We investigate how \emph{missing vision modality} affects confidence reliability in VLMs. Across the following three studies, we show that (i) removing the image modality while largely preserving semantic content in the textual description substantially degrades performance and, more prominently, induces \textbf{severe miscalibration}; (ii) under text-only queries, VLMs do \emph{not} reliably fall back to their language backbones, often exhibiting worse accuracy and calibration than the backbone LLM itself; and (iii) even on intrinsically text-only question answering task, injecting auxiliary visual inputs can improve both accuracy and calibration. Our experiments are conducted on the visual question answering (VQA) dataset \textsc{VQA-v1}~\citep{antol2015vqa}, where each example consists of an image–question pair with a ground-truth answer. Unless otherwise specified, we refer to \textsc{LLaVA} as \textsc{LLaVA-1.5-7B} and \textsc{Vicuna} as \textsc{Vicuna-1.5-7B} throughout this paper.

\subsection{On Text-Only Miscalibration under Image-to-Text Description}
\label{sec:finding_caption_and_regen}

\paragraph{Image-to-text description substitution.}
We first probe missing-modality behavior by \emph{contrasting} standard VQA inference with full visual evidence against a caption-only variant in which the image is replaced by a high-fidelity textual description produced by an image captioning model.
Specifically, for each VQA instance, we evaluate LLaVA under two conditions:
\textbf{(i) Image+Question}, where the original image and the question are provided as usual, and
\textbf{(ii) Description+Question}, where the image is substituted with its text description while keeping the question unchanged. 
This substitution protocol is inspired by the evaluation setup of \citet{chen2025unveiling}, which compares paired VQA inference against a caption-substituted (text-only) variant. We further extend this setting by re-introducing a visual modality signal via text-to-image regeneration from the same description. Details for description generation are provided in Appendix~\ref{app:description generation}.

\begin{figure*}[t]
    \centering
    \begin{minipage}{0.3\textwidth}
        \centering
        \includegraphics[width=\linewidth]{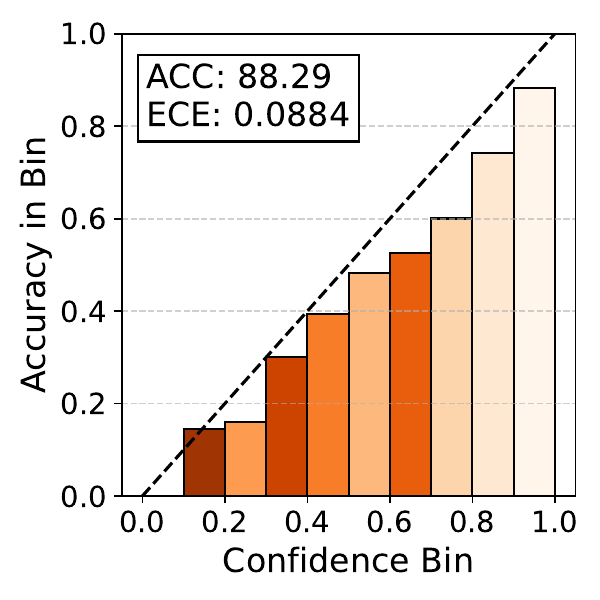}
        \small (a) Image
    \end{minipage}\hfill
    \begin{minipage}{0.3\textwidth}
        \centering
        \includegraphics[width=\linewidth]{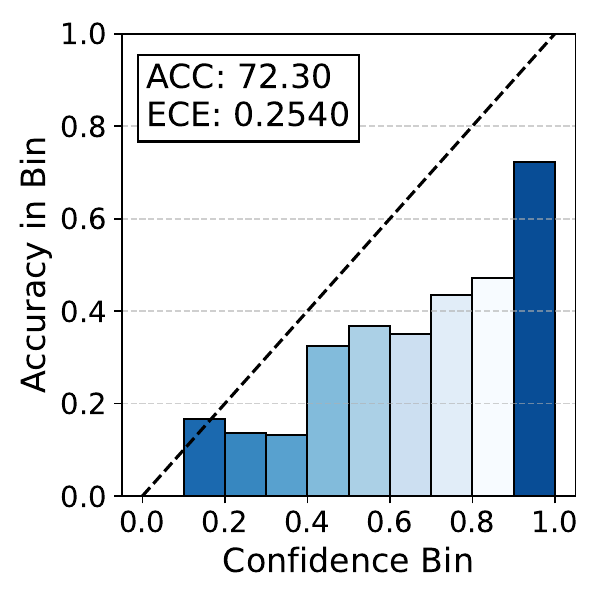}
        \small (b) Text description
    \end{minipage}\hfill
    \begin{minipage}{0.3\textwidth}
        \centering
        \includegraphics[width=\linewidth]{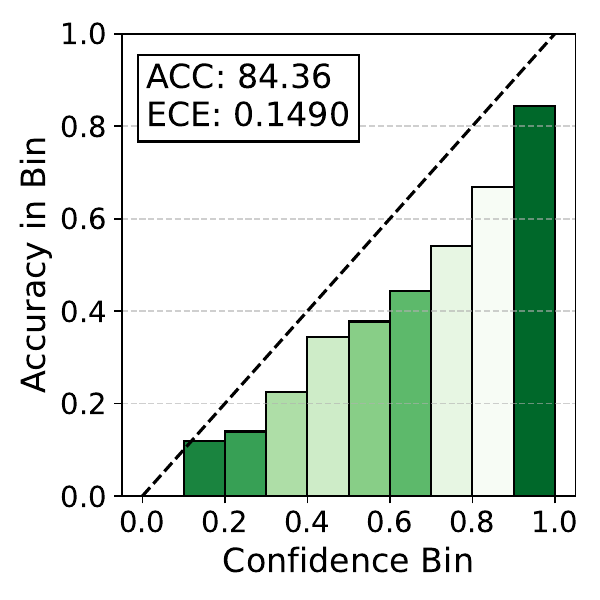}
        \small (c) Re-generated image
    \end{minipage}
    \caption{\textbf{Reliability diagrams on \textsc{VQA-v1} (4,500 test samples) using \textsc{LLaVA}}. Replacing the image with a text description induces severe over-confidence, while re-introducing an image-modality signal via text-to-image regeneration partially restores calibration. \textit{Bar intensity indicates the number of samples in each confidence bin.}}
    \label{fig:reliability_all}
    \vspace{-6mm}
\end{figure*}

According to the results in Figure~\ref{fig:reliability_all}, relative to the Image+Question baseline, the Description+Question setting shows a drop in accuracy (88.29\% $\rightarrow$ 72.30\%). Nevertheless, performance remains well above chance, suggesting that the description preserves a meaningful portion of task-relevant semantic information. 
In contrast, confidence reliability deteriorates sharply: the expected calibration error (ECE)~\citep{ece} increases from 0.0884 to 0.2540, indicating systematic overconfidence across most confidence bins, where predicted confidence remains high despite the pronounced reduction in empirical accuracy. 

\paragraph{Re-generated images from descriptions.}
The degradation under Description+Question could, in principle, arise from \emph{information loss} introduced when converting images into text descriptions. To examine this possibility, we generate an image from the description using a diffusion-based text-to-image model (Stable Diffusion v2.1 base; \texttt{stabilityai/stable-diffusion-2-1-base}) with 50 inference steps, and evaluate \textsc{LLaVA} on the VQA task by providing the re-generated image together with the original question.
We find that reintroducing a visual input substantially restores model behavior.
Compared to the Description+Question condition, the re-generated image leads to a clear recovery in accuracy (ACC: 72.30 $\rightarrow$ 84.36) and calibration (ECE: 0.2540 $\rightarrow$ 0.1490), with the reliability curve moving closer to the diagonal (Figure~\ref{fig:reliability_all}(c)).
Taken together, these results suggest that the severe overconfidence in the caption-only setting is not fully attributable to information loss from image-to-text conversion. While the description may omit fine-grained visual details, the fact that providing even a re-generated image substantially restores both accuracy and calibration indicates that the presence of a visual-modality signal, together with the corresponding multimodal processing pathway, plays an important role in reliable confidence estimation. 

\subsection{VLMs Do Not Fall Back to Their Language Backbones under Text-Only Inputs}
\label{sec:finding_backbone_gap}

We next ask whether a VLM queried with \emph{text-only} inputs behaves like its underlying language backbone, as one might naturally expect when no image is provided.

\paragraph{VLM vs.\ LLM backbone under text-only prompting.}
We compare \textsc{Vicuna} (a text-native LLM) and \textsc{LLaVA} (a VLM built on a \textsc{Vicuna-1.5-7B} backbone) on a text-only evaluation suite consisting of \textsc{MMLU}~\citep{hendrycks2020measuring} and \textsc{ARC}~\citep{clark2018think} (ARC-Easy and ARC-Challenge), distinct from \textsc{VQA-v1}.
Both models receive exactly the same textual input (question, multiple-choice options, and instructions), and we match decoding settings across models.
If a VLM reliably ``falls back'' to its language backbone when no image is provided, its accuracy and calibration should be comparable to those of LLM.
Contrary to this expectation, \textsc{LLaVA} consistently performs worse than \textsc{Vicuna} on text-only queries. As shown in Table~\ref{tab:vicuna_llava_calibration} (aggregated over MMLU+ARC), LLaVA achieves lower accuracy than Vicuna (ACC@logits: 39.77 vs.\ 53.87) and exhibits dramatically worse calibration (ECE@logits: 0.4202 vs.\ 0.2011).

\paragraph{Not an artifact of logit-based confidence.}
Our main analysis uses the maximum softmax probability (MSP) confidence in answer-choice logits~\citep{hendrycks2016baseline} and evaluates calibration with ECE~\citep{guo2017calibration}. However, the observed gap is not specific to this estimator. We repeat the evaluation using alternative uncertainty estimators, including \emph{semantic entropy}~\citep{kuhn2023semantic} and a \emph{top-$k$ margin} confidence signal~\citep{wang2024chain}, and observe the same qualitative trend: \textsc{LLaVA} remains significantly more over-confident than \textsc{Vicuna} under text-only inputs. This indicates that the text-only failure mode arises from the VLM inference dynamics under missing-modality conditions, rather than from a particular choice of confidence estimator.

\paragraph{Additional model family.}
To verify that this missing-modality behavior is not unique to the \textsc{Vicuna/LLaVA} family, we also examine a \textsc{Qwen}-based VLM family under the same \emph{text-only} prompting setup using our primary MSP confidence estimator. This sanity check reveals the same qualitative pattern of overconfidence under missing vision modality. We report the corresponding ACC/ECE and reliability diagrams in Appendix~\ref{app:qwen_additional}.

\begin{table}[!t]
\centering
\caption{Text-only evaluation on 
\textsc{MMLU+ARC (ARC-Easy/Challenge)}: comparison of Vicuna vs.\ LLaVA across multiple uncertainty estimators.}
\vspace{1mm}
\renewcommand{\arraystretch}{1.15}
\setlength{\tabcolsep}{5pt}
\resizebox{\columnwidth}{!}{%
\begin{tabular}{l cccccc}
\toprule
\textbf{Model} &
\textbf{ACC@logits} $\uparrow$ &
\textbf{ECE@logits} $\downarrow$ &
\textbf{ACC@entropy} $\uparrow$ &
\textbf{ECE@entropy} $\downarrow$ &
\textbf{ACC@topk} $\uparrow$ &
\textbf{ECE@topk} $\downarrow$ \\
\midrule
Vicuna-1.5-7B & 53.87 & 0.2011 &  52.85 &  0.0709 & 53.02 & 0.0920 \\
LLaVA-1.5-7B  & 39.77 & 0.4202 &  39.42 &  0.1198 & 39.15 & 0.1615\\
\bottomrule
\end{tabular}%
}
\label{tab:vicuna_llava_calibration}
\end{table}

\subsection{Improving Text-Only QA by Injecting Diffusion-Generated Visual Inputs}
\label{sec:finding_text_task_with_generated_images}

Finally, we consider a strictly \emph{text-only} regime where no ground-truth image exists. Our goal is to validate whether providing an auxiliary visual input can improve a VLM's decisions and confidence reliability in text-only deployment.

\paragraph{Auxiliary visual augmentation.}
Given a text-only QA instance, we prompt a text-to-image generator (Stable Diffusion) with the question and instruct it to generate an image that could be helpful for solving the problem.
We then feed this generated image to the VLM alongside the original question, treating it as the model's visual input. Prompt templates and generation hyperparameters for diffusion-based auxiliary images are provided in Appendix~\ref{app:helpful_image_prompt}. Empirically, adding diffusion-generated images yields a modest improvement over the text-only VLM baseline (ACC: {39.77}\% $\rightarrow$ {40.05}\%; ECE: {0.4202} $\rightarrow$ {0.3952}). While the improvement is small, this observation provides preliminary evidence that introducing a visual modality signal may influence VLM confidence behavior even in text-only QA settings.

Despite its effectiveness, diffusion-based generation is computationally expensive and impractical as a general-purpose inference-time augmentation.
Motivated by this observation, the remainder of the paper develops a more efficient alternative that retains the benefit of ``having a visual modality'' without explicitly sampling pixels. Instead, we inject a lightweight \emph{latent imagination} signal alongside text-only inputs, aiming to recover the calibration gains observed here while avoiding the cost of diffusion sampling.

\section{Method}
\label{method}

\begin{figure*}[t]
    \centering
    \begin{minipage}{0.49\textwidth}
        \centering
        \includegraphics[width=\linewidth]{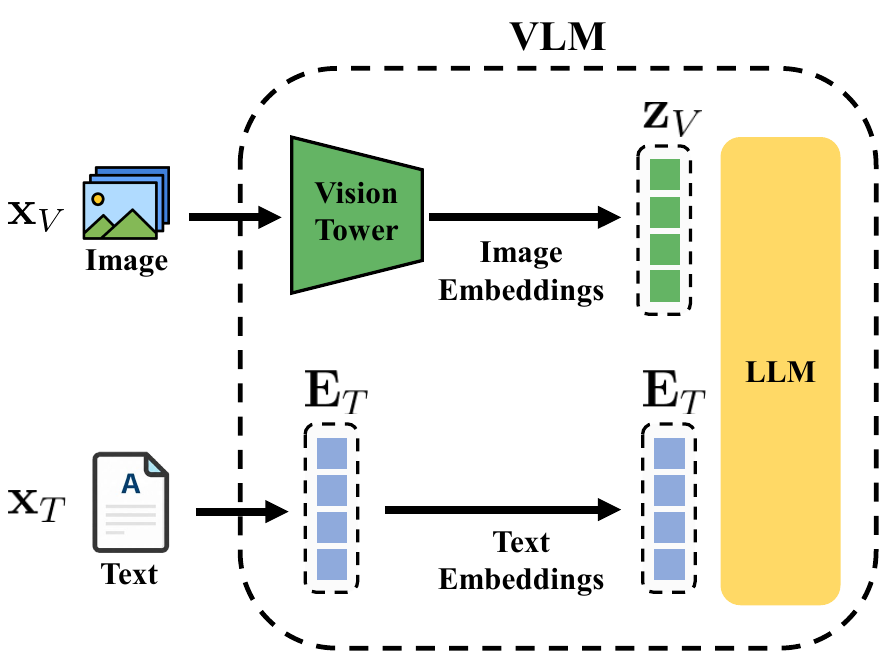}
        \small (a) Standard VLM
    \end{minipage}\hfill
    \begin{minipage}{0.49\textwidth}
        \centering
        \includegraphics[width=\linewidth]{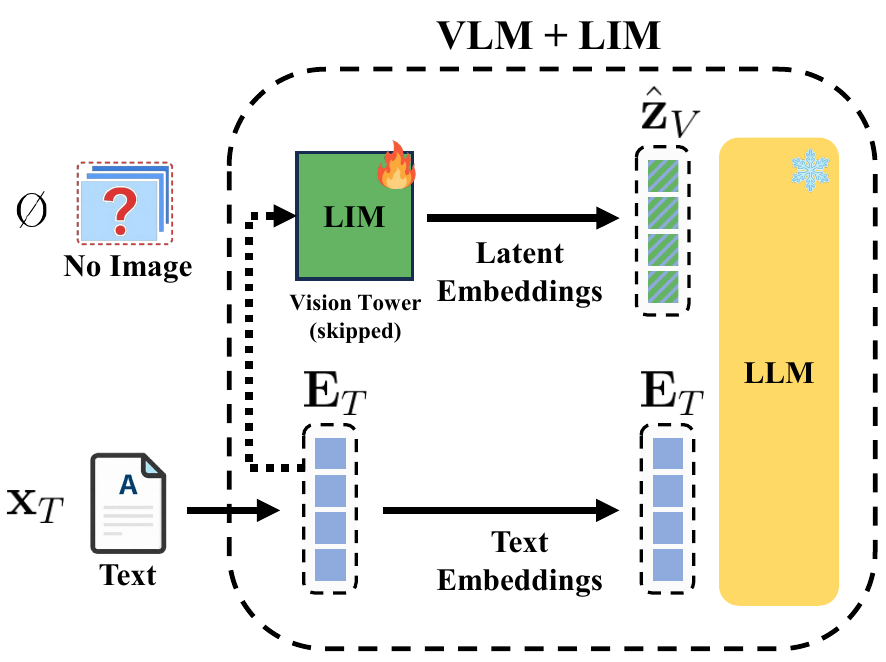}
        \small (b) VLM with LIM
    \end{minipage}
    \caption{\textbf{Approach overview.} \textbf{(a)} Standard VLM inference with image and text inputs. \textbf{(b)} Text-only inference with LIM, which predicts and injects latent embeddings.}
    \label{fig:method_overview}
\end{figure*}

\subsection{Problem Formulation}
Consider a multimodal dataset $\mathcal{D} = \{(\mathbf{x}_T, \mathbf{x}_V, y)\}$, where $\mathbf{x}_T$ represents the textual input (e.g., a question), $\mathbf{x}_V$ represents the visual input (e.g., an image), and $y$ is the target answer. Standard VLMs are trained to maximize $P(y \mid \mathbf{x}_T, \mathbf{x}_V)$. However, in real-world deployment, the visual modality $\mathbf{x}_V$ can be missing in two distinct scenarios. First, it may be physically unavailable (the task is inherently text-only), and no corresponding image exists. Second, even when visual information is conceptually relevant, it may be operationally absent, as the model is queried with text-only inputs due to deployment constraints. In both cases, the model is forced to infer $y$ solely from $\mathbf{x}_T$. We denote these missing-modality scenarios as $P(y \mid \mathbf{x}_T, \emptyset)$. Our empirical findings (\cref{findings}) suggest that these absences lead to miscalibration. We aim to recover calibration by predicting a latent visual representation $\hat{\mathbf{z}}_V$ such that $P(y \mid \mathbf{x}_T, \hat{\mathbf{z}}_V)$ approximates the reliability of the full-modality predictive distribution.

\subsection{Latent Imagination Module}
\begin{wrapfigure}{r}{0.42\columnwidth}
    \centering
    \vspace{-13mm}
    \includegraphics[width=\linewidth]{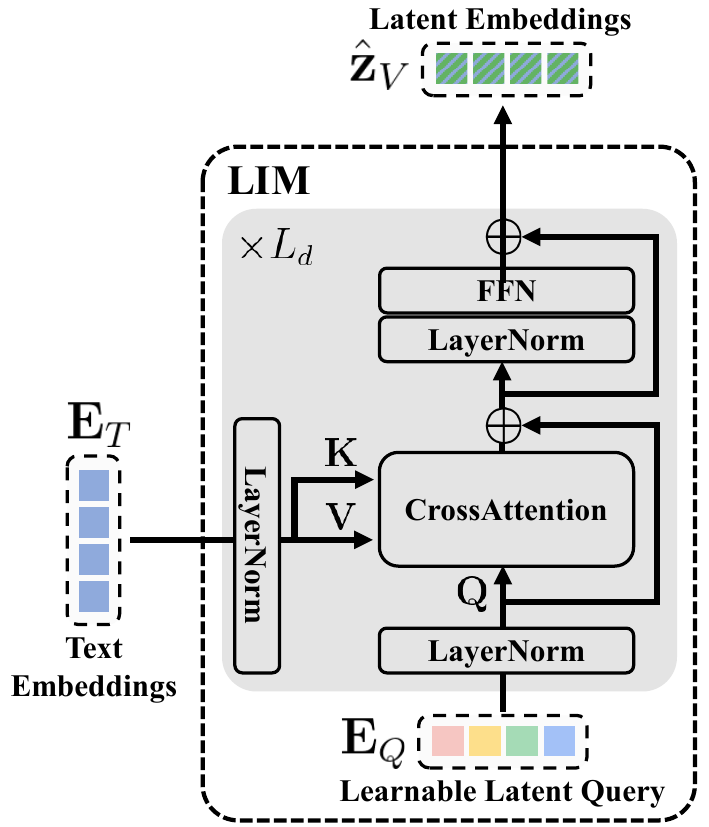}
    \caption{Architecture of the Latent Imagination Module.}
    \label{fig:lim_details}
    \vspace{-4mm}
\end{wrapfigure}
To infer the missing visual information, we introduce the \textbf{Latent Imagination Module (LIM)}. Unlike generative approaches that synthesize raw pixels (e.g., diffusion models), LIM predicts high-dimensional feature embeddings directly compatible with the VLM's vision tower, which comprises an image encoder followed by a projection layer. This bypasses the computational overhead of pixel generation and focuses on task-relevant semantics (Figure~\ref{fig:method_overview}).

\noindent\textbf{Learnable latent queries.}
We formulate the imagination process as a query-based decoding task. Let $\mathbf{E}_T \in \mathbb{R}^{L \times D}$ be the sequence of text embeddings extracted from the VLM's text embedding layer, where $L$ is the sequence length and $D$ is the embedding dimension. We initialize a set of $N$ learnable latent queries $\mathbf{E}_Q \in \mathbb{R}^{N \times D}$, where $N$ corresponds to the number of visual tokens (patches) the VLM expects. These queries serve as placeholders for the missing visual concepts, to be populated with information from the text.

\noindent\textbf{Cross-attentive imputation.}
The core of our module is a stack of Cross-Attention blocks (Figure~\ref{fig:lim_details}) that progressively refine the queries using the textual context. We incorporate sinusoidal positional encodings $\mathbf{P}_Q$ and $\mathbf{P}_T$ to capture the structural order. The process is defined as follows:

\begin{equation}
    \mathbf{Q}^{(0)} = \mathbf{E}_Q + \mathbf{P}_Q, \quad \mathbf{H}_T = \mathbf{E}_T + \mathbf{P}_T
\end{equation}

For each layer $l \in \{1, \dots, L_d\}$, where $L_d$ is the number of cross-attention layers:
\begin{align}
    \hat{\mathbf{Q}}^{(l)} &= \text{CrossAttention}(\text{LN}(\mathbf{Q}^{(l-1)}), \text{LN}(\mathbf{H}_T), \text{LN}(\mathbf{H}_T)) + \mathbf{Q}^{(l-1)} \\ \label{eq:crossattn}
    \mathbf{Q}^{(l)} &= \text{FFN}(\text{LN}(\hat{\mathbf{Q}}^{(l)})) + \hat{\mathbf{Q}}^{(l)}
\end{align}
where $\text{LN}(\cdot)$ denotes Layer Normalization. We adopt this pre-layer normalization (Pre-LN) configuration to ensure training stability and architectural consistency with the underlying VLM backbone.

Here, $\text{CrossAttention}(\mathbf{Q}, \mathbf{K}, \mathbf{V})$ computes the standard scaled dot-product attention:
\begin{equation}
    \text{Attention}(\mathbf{Q}, \mathbf{K}, \mathbf{V}) = \text{softmax}\left(\frac{\mathbf{Q} \mathbf{K}^\top}{\sqrt{D}}\right) \mathbf{V}
\end{equation}
The final output $\hat{\mathbf{z}}_V = \mathbf{Q}^{(L_d)}$ represents the synthesized visual features, which are then injected into the VLM's input sequence in place of the missing image tokens.

\subsection{Optimization via Task-Oriented Grounding}
\label{sec:optimization}

A critical design choice in our framework is the training objective. We optimize the LIM end-to-end using the downstream task loss rather than a latent reconstruction loss.

\noindent\textbf{Task-oriented objective.}
We train LIM to minimize the negative log-likelihood of the correct answer $y$ given the text $\mathbf{x}_T$ and the envisioned visual features $\hat{\mathbf{z}}_V$:
\vspace{-1pt}
\begin{equation}
\begin{aligned}
    \theta_{\text{LIM}} \leftarrow
\arg\min_{\theta_{\text{LIM}}}\;
\mathbb{E}_{(\mathbf{x}_T,y)\sim \mathcal{D}}
\big[-\log P(y\mid \mathbf{x}_T,\hat{\mathbf{z}}_V)\big]
+\lambda \|\theta_{\text{LIM}}\|_2^2.
\end{aligned}
\end{equation}
\vspace{-2pt}
During the training phase, the parameters of the VLM ($\theta_{\text{VLM}}$) remain frozen to preserve its original multimodal alignment, while only the LIM parameters ($\theta_{\text{LIM}}$) are updated.

One might consider training the LIM to minimize the distance (e.g. mean squared error) between the predicted embeddings $\hat{\mathbf{z}}_V$ and the ground-truth image embeddings $\mathbf{z}_V$ (extracted from the oracle vision tower). However, we argue that this approach is suboptimal for textual calibration due to the \textit{one-to-many} nature of the problem.
A single text description (e.g., ``a dog sitting on a bench'') can correspond to infinite visual realizations (different breeds, poses, backgrounds). Enforcing a distance-based loss would compel the model to predict the \textit{mean} of all possible visual features, often resulting in ``blurry'' or overly smoothed representations that lack semantic distinctiveness.
By contrast, our task-oriented objective relaxes this strict constraint. It allows the model to synthesize \textit{any} valid visual representation that is semantically sufficient to solve the task. This "sufficient imagination" serves as a better regularizer for the VLM's confidence, as it aligns the latent features with the decision boundary of the task rather than pixel-perfect statistics of a specific image instance.

\section{Experiments}
\label{sec:experiments}

In this section, we evaluate whether the proposed LIM improves both \emph{accuracy} and \emph{confidence reliability} in missing-modality regimes.
Unlike pixel-level generation, LIM directly predicts the latent embeddings expected by the VLM and injects them in place of missing image features (Section~\ref{method}).
Across three experiment families, we show that: 
(i) LIM substantially improves text-only QA performance and calibration; 
(ii) the gains also transfer to \textbf{unseen} text-only tasks; and 
(iii) LIM improves robustness when images are \textbf{partially missing in originally paired image-text tasks}.

\paragraph{Models and baselines.}
We use \textsc{LLaVA-1.5-7B} as our base model, comparing \textbf{(a) LLaVA (text-only)}, where the visual modality is absent, against \textbf{(b) LLaVA+LIM}, where LIM injects $\hat{\mathbf{z}}_V$ as the missing visual signal.
Unless otherwise specified, the VLM backbone remains frozen and only LIM is trained. Thus, our approach does not steer the VLM toward task-specific behavior via finetuning, but instead improves text-only inference by supplying missing visual signals in latent space.

\paragraph{Training dataset.}
LIM is trained on an aggregated \textsc{MMLU+ARC} text-only suite.
Specifically, we combine the \textsc{MMLU} \texttt{auxiliary\_train} split with the \textsc{ARC} train splits (\textsc{ARC-Easy} and \textsc{ARC-Challenge}), then randomly sample 2,000 instances to construct the LIM training set. All samples are formatted as text-only multiple-choice QA prompts following the evaluation protocol in Section~\ref{sec:finding_backbone_gap}, and the corresponding ground-truth answer choice is used as supervision. Unless otherwise specified, we train LIM only (freezing the VLM backbone) on this 2,000-example subset.
We use AdamW with learning rate $10^{-4}$, weight decay $0.01$, batch size $8$, and train for $3$ epochs.

\subsection{Recovering Text-Only Calibration with LIM}
\label{sec:exp_text_only_main}

\begin{wraptable}{r}{0.442\columnwidth}
\centering
\vspace{-12pt}
\caption{\textbf{Inference-time overhead: diffusion pixels vs.\ LIM (BS=1).}
End-to-end latency and peak VRAM per sample.}
\label{tab:latency_sd_vs_lim}
\renewcommand{\arraystretch}{1.05}
\vspace{-9pt}
\adjustbox{max width=\linewidth}{%
\begin{tabular}{lccc}
\toprule
\textbf{Method} & \textbf{Latency} $\downarrow$ & \textbf{VRAM} & \textbf{TFLOPs}\footnotemark{} $\downarrow$\\
\midrule
SD + LLaVA & 2.52s & 9.64GB & 16.98\\
LIM + LLaVA & 0.205s & 9.44GB & 0.20\\
\midrule
Speedup & 12.26$\times$ & --  & 85.42$\times$\\
\bottomrule
\end{tabular}}
\vspace{-10pt}
\end{wraptable}
\footnotetext{TFLOPs are measured using \texttt{fvcore}~(\url{https://github.com/facebookresearch/fvcore}).}

We first report in-domain results on the text-only suite used to train LIM, i.e., the \textsc{MMLU+ARC} evaluation suite from Section~\ref{sec:finding_backbone_gap}.
Given a textual input, LIM predicts latent embeddings $\hat{\mathbf{z}}_V$ in the backbone language model's embedding space and injects them as a surrogate visual signal; all language inputs and decoding settings are kept identical to the text-only baseline (i.e., \textsc{LLaVA} without visual input).

\begin{figure*}[t]
    \centering
    \begin{subfigure}{0.25\textwidth}
        \centering
        \includegraphics[width=\linewidth]{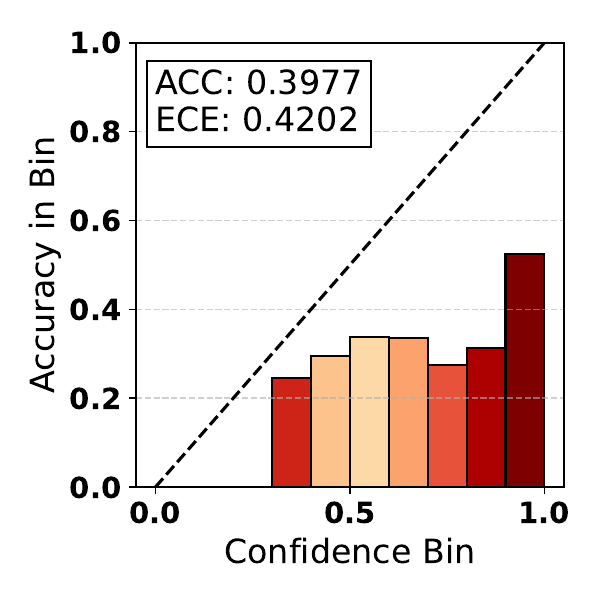}
        \caption{Baseline}
        \label{fig:tomr_a}
    \end{subfigure}\hfill
    \begin{subfigure}{0.25\textwidth}
        \centering
        \includegraphics[width=\linewidth]{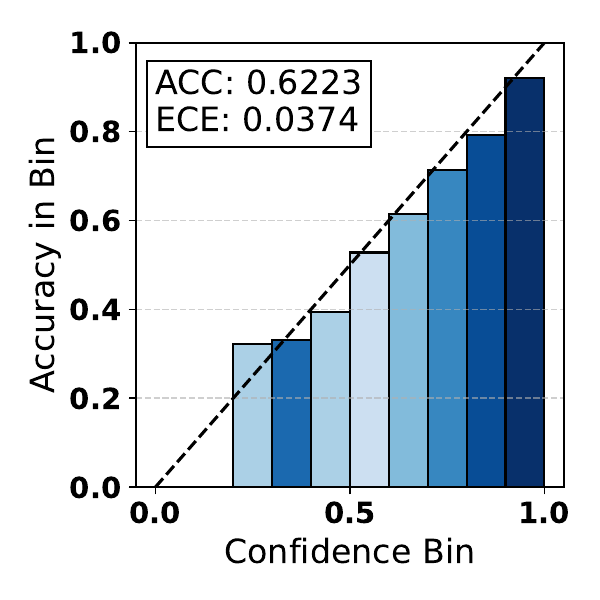}
        \caption{Ours}
        \label{fig:tomr_b}
    \end{subfigure}\hfill
    \begin{subfigure}{0.25\textwidth}
        \centering
        \includegraphics[width=\linewidth]{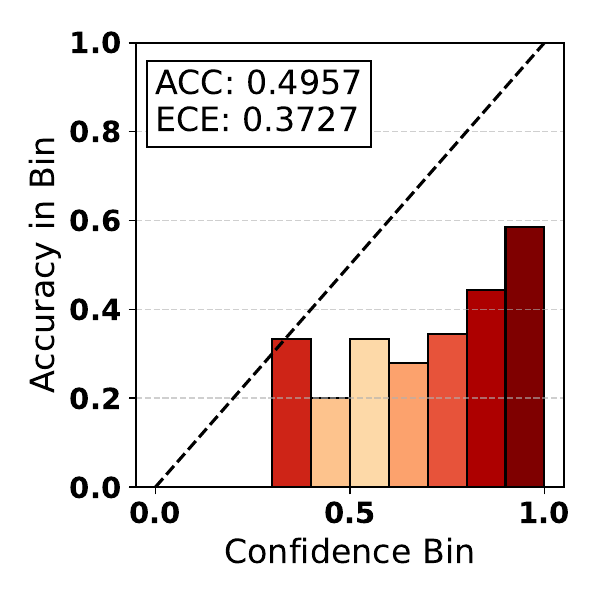}
        \caption{Baseline (unseen)}
        \label{fig:tomr_c}
    \end{subfigure}\hfill
    \begin{subfigure}{0.25\textwidth}
        \centering
        \includegraphics[width=\linewidth]{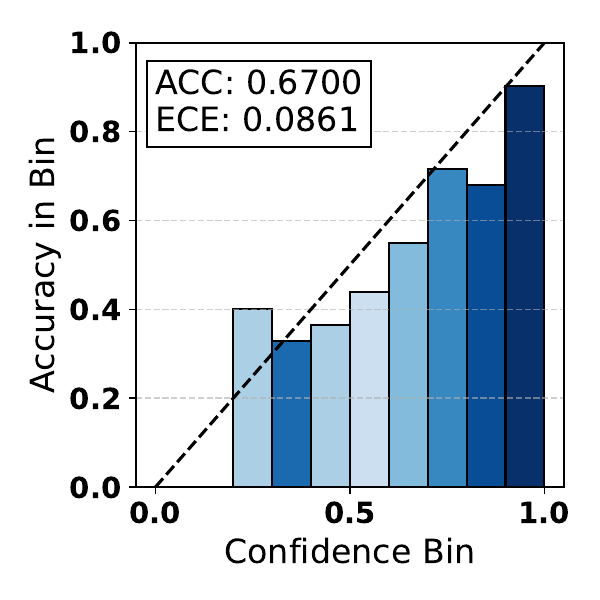}
        \caption{Ours (unseen)}
        \label{fig:tomr_d}
    \end{subfigure}
    \caption{Reliability diagrams for text-only QA.
(\subref{fig:tomr_a}) LLaVA (text-only) and (\subref{fig:tomr_b}) LLaVA+LIM on the in-domain MMLU+ARC suite;
(\subref{fig:tomr_c}) LLaVA (text-only) and (\subref{fig:tomr_d}) LLaVA+LIM on eight unseen text-only benchmarks (aggregated).
Injecting LIM consistently reduces over-confidence and moves the reliability curve closer to the diagonal.}
    \label{fig:text_only_main_rel}
\end{figure*}

Table~\ref{tab:text_only_main} shows that LIM improves both accuracy and calibration by a large margin.
ECE drops from {0.4202} to {0.0374}, while accuracy increases from {39.77\%} to {62.23\%}.
Figures~\ref{fig:tomr_a} and \ref{fig:tomr_b} corroborate this trend: \textbf{LLaVA+LIM} yields a reliability curve substantially closer to the diagonal, indicating reduced over-confidence.

\paragraph{Comparison with Temperature Scaling.}
We also compare LIM against temperature scaling~\citep{guo2017calibration}, a widely used post-hoc calibration method.
Following standard practice, we optimize a single temperature parameter on a held-out validation split of 500 examples using the negative log-likelihood objective, and apply the calibrated temperature to the same test set.
Temperature scaling substantially reduces ECE for the text-only baseline (from 0.42 to 0.095) while leaving accuracy unchanged.
In contrast, LIM achieves stronger calibration than temperature scaling (ECE = 0.037 vs.\ 0.095) while simultaneously improving accuracy.
Applying temperature scaling on top of LIM yields only marginal changes (ECE = 0.037 $\rightarrow$ 0.0495), suggesting that LIM already resolves much of the miscalibration at the representation level.

\paragraph{Efficiency of pixel-space generation vs.\ latent token prediction.}
Diffusion-based auxiliary images are effective (Section~\ref{sec:finding_text_task_with_generated_images}) but incur substantial inference overhead due to iterative pixel sampling.
Table~\ref{tab:latency_sd_vs_lim} shows that, under identical decoding settings and batch size $1$, LIM reduces end-to-end latency from 2.52 to 0.205 s/sample (12.26$\times$) with similar peak VRAM (9.64 vs.\ 9.44 GB).
In the diffusion baseline, image generation dominates runtime (2.38 s/sample; \(\sim\)94\% of total).

\subsection{Generalization to Unseen Text-Only Tasks}
\label{sec:exp_text_only_unseen}

We next evaluate whether LIM transfers beyond the training distribution.
Concretely, we test the same trained LIM (no further tuning) on eight unseen text-only benchmarks that are not used during LIM training or model selection: \textsc{SST-2}, \textsc{CoLA}, \textsc{AG News}, \textsc{MRPC}, \textsc{Vitamin-C}, \textsc{LogiQA}, \textsc{CommonsenseQA}, and \textsc{QASC}. These datasets cover heterogeneous NLP skills, including sentiment analysis and topic classification (\textsc{SST-2}~\citep{socher2013recursive}, \textsc{AG News}~\citep{zhang2015character}),
grammatical acceptability (\textsc{CoLA}~\citep{warstadt2019neural}),
sentence-pair semantic equivalence (\textsc{MRPC}~\citep{dolan2005automatically}),
factual correction and evidence-based consistency (\textsc{Vitamin-C}~\citep{schuster2021get}),
logical reasoning (\textsc{LogiQA}~\citep{liu2020logiqa}),
commonsense reasoning (\textsc{CommonsenseQA}~\citep{talmor2019commonsenseqa}),
and science QA (\textsc{QASC}~\citep{khot2020qasc}).
We keep the inference pipeline and decoding settings identical to Section~\ref{sec:exp_text_only_main}. Table~\ref{tab:text_only_main} reports both in-domain and per-task unseen results. On most benchmarks, \textbf{LLaVA+LIM} improves calibration compared to the text-only baseline, and often improves accuracy as well, indicating that LIM learns a transferable missing-modality correction rather than overfitting to MMLU+ARC (in-domain).
\vspace{2em}

\begin{table}[!t]
\centering
\small
\caption{\textbf{In-domain and unseen text-only benchmarks.}
ACC and ECE for the text-only baseline vs.\ LLaVA+LIM.
MMLU+ARC is in-domain (used to train LIM); all others are unseen.}
\vspace{-1mm}
\renewcommand{\arraystretch}{1}
\setlength{\tabcolsep}{9pt}

\begin{tabular}{l|cc|cc}
\toprule
\multirow{2}{*}{\textbf{Benchmark}} &
\multicolumn{2}{c|}{\textbf{Baseline}} &
\multicolumn{2}{c}{\textbf{LLaVA+LIM}} \\
\cmidrule(lr){2-3}\cmidrule(lr){4-5}
 & \textbf{ACC (\%)} $\uparrow$ & \textbf{ECE} $\downarrow$ &
   \textbf{ACC (\%)} $\uparrow$ ($\Delta$) & \textbf{ECE} $\downarrow$ ($\Delta$) \\
\midrule
\textbf{MMLU+ARC (in-domain)} & 39.77 & 0.4202 & \gain{62.23}{(+22.46)} & \gain{0.0374}{(-0.3828)} \\
\midrule
\textsc{SST-2}          & 96.50 & 0.0581 & \same{96.50}{(+0.00)} & \loss{0.0595}{(+0.0014)} \\
\textsc{CoLA}           & 32.50 & 0.6119 & \gain{68.00}{(+35.50)} & \gain{0.1047}{(-0.5072)} \\
\textsc{AG News}        & 61.50 & 0.2018 & \gain{79.50}{(+18.00)} & \gain{0.0594}{(-0.1424)} \\
\textsc{MRPC}           & 34.00 & 0.5354 & \gain{66.00}{(+32.00)} & \gain{0.2305}{(-0.3049)} \\
\textsc{Vitamin-C}      & 51.00 & 0.4534 & \gain{60.00}{(+9.00)} & \gain{0.1016}{(-0.3518)} \\
\textsc{LogiQA}         & 28.00 & 0.5313 & \gain{40.00}{(+12.00)} & \gain{0.1360}{(-0.3953)} \\
\textsc{CommonsenseQA}  & 48.00 & 0.3343 & \gain{75.00}{(+27.00)} & \gain{0.0831}{(-0.2512)} \\
\textsc{QASC}           & 39.00 & 0.3442 & \gain{43.00}{(+4.00)} & \gain{0.2262}{(-0.1180)} \\
\midrule
\textbf{Average (unseen)} & 48.81 & 0.3838 & \gain{66.00}{(+17.19)} & \gain{0.1251}{(-0.2587)} \\
\bottomrule
\end{tabular}%
\vspace{-2mm}
\label{tab:text_only_main}
\end{table}

\subsection{Robustness to Missing Images in Paired vision-language Tasks}
\label{sec:exp_missing_images_paired}

We further evaluate LIM in a \emph{paired} vision-language setting, where each instance originally contains both image and text, but images may be missing at inference time.
We use the \textbf{ScienceQA}~\citep{lu2022learn} test split and restrict evaluation to the \emph{image-subset} (2,017 instances with available images), then simulate missing-modality by dropping images with probability $p \in \{0.25, 0.50, 0.75, 1.00\}$.
For instances whose images are not dropped, we run standard LLaVA inference with the original image.
For dropped instances, we compare two missing-modality strategies:
\textbf{(i) drop-and-text-only}, which removes the image and then queries LLaVA with text-only,
and \textbf{(ii) drop-and-LIM}, which injects LIM-predicted latent embeddings $\hat{\mathbf{z}}_V$ in place of the missing image features (injection at position (B); Section~\ref{method}).
All other prompting and decoding settings are kept identical.
\vspace{-0.5em}

\begin{figure*}[t]
    \centering

\begin{subfigure}{0.65\textwidth}
    \centering
    \begin{subfigure}{0.48\linewidth}
        \centering
        \includegraphics[width=\linewidth]{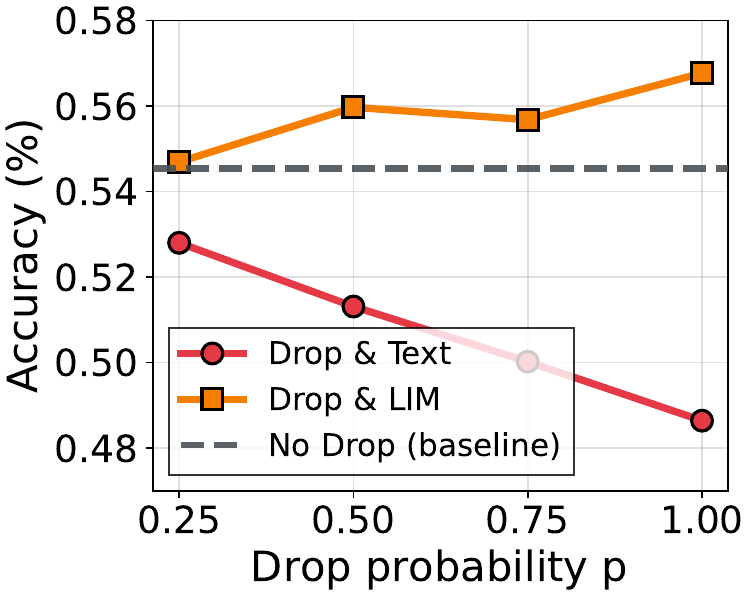}
        \label{fig:scienceqa_drop_acc}
    \end{subfigure}
    \begin{subfigure}{0.48\linewidth}
        \centering
        \includegraphics[width=\linewidth]{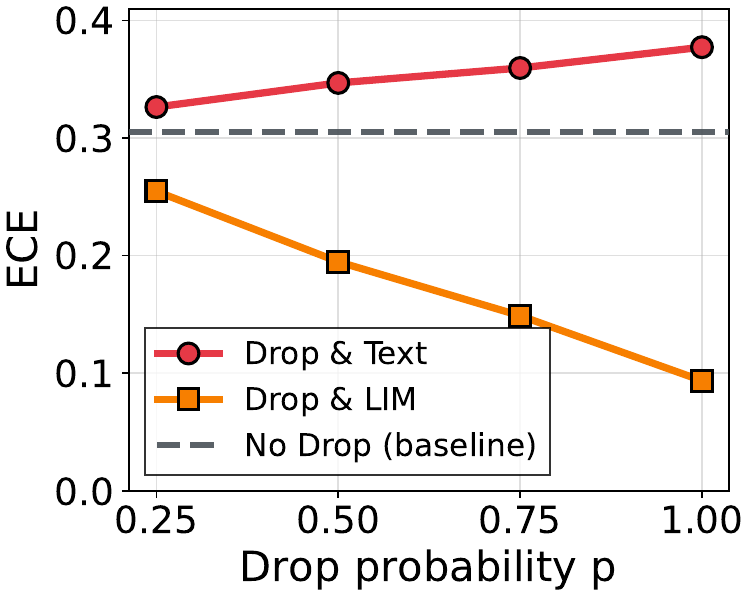}
        \label{fig:scienceqa_drop_ece}
    \end{subfigure}
    \vspace{-5mm}
    \caption{Missing-modality robustness.}
    \label{fig:scienceqa_drop_main}
\end{subfigure}
\begin{subfigure}{0.34\textwidth}
    \centering
    \includegraphics[width=\linewidth]{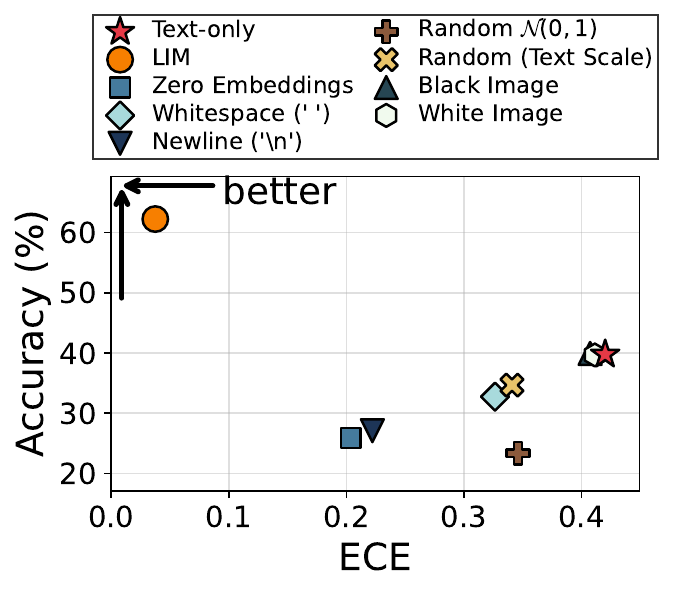}
    \vspace{-5mm}
    \caption{Visual substitute ablations.}
    \label{fig:ablation_randomness}
\end{subfigure}
\vspace{-5mm}
\caption{\textbf{(a) ScienceQA (image-subset): robustness to missing images.} We evaluate on 2,017 ScienceQA test instances with images and simulate missing-modality by dropping images with probability $p$. For dropped instances, \textit{drop-and-text-only} queries LLaVA with text-only, while \textit{drop-and-LIM} injects LIM-predicted latent embeddings in place of missing image features. 
\textbf{(b) Ablations on filling missing visual signal with arbitrary substitutes.} We replace missing visual-token slots with Zero vectors, repeated Whitespace/Newline token embeddings, or Random vectors (scale-matched or $\mathcal{N}(0,1)$), and compare against LIM. We additionally test Blank image inputs (all-black/all-white) by providing a content-free image with the same text prompt.
} 
\vspace{-2mm}
\end{figure*}

\paragraph{Results.}
Figure~\ref{fig:scienceqa_drop_main} summarizes accuracy and ECE as a function of the missing rate $p$.
Under \textbf{drop-and-text-only}, performance degrades monotonically as more images are removed (ACC: $54.54 \rightarrow 48.64$), and calibration deteriorates substantially (ECE: $0.3054 \rightarrow 0.3773$), consistent with the over-confidence failure mode observed under missing-modality.
In contrast, \textbf{drop-and-LIM} remains robust as $p$ increases, maintaining (and slightly improving) accuracy while markedly improving reliability.
At $p{=}1.0$ (all images dropped), \textbf{drop-and-LIM} achieves comparable accuracy to the full-image baseline and substantially better calibration (ACC: $56.77$ vs.\ $54.54$, ECE: $0.0936$ vs.\ $0.3054$), suggesting that LIM provides a task-oriented ``imagination'' signal that the VLM can leverage more reliably, improving both predictive correctness and confidence reliability.
\vspace{-0.5em}

Overall, these results show that LIM is effective not only for intrinsically text-only QA, but also for \emph{paired} vision-language tasks under partial image unavailability, substantially reducing over-confidence while preserving (or improving) task accuracy.
\vspace{-0.5em}

\subsection{Ablation Study: Are Improvements Driven by Arbitrary Visual Tokens?}
\label{sec:ablation}
\vspace{-0.5em}

A natural concern is that the gains from LIM might come from simply \emph{filling the missing visual-token slots} with \emph{any} embeddings, rather than from meaningful latent imagination.
To test this hypothesis, we conduct a set of controlled ablations where we replace the missing image tokens with alternative embeddings that are \emph{not} produced by LIM.
Importantly, all ablation embeddings are injected at the same position as LIM, and the VLM backbone, prompting, and decoding settings are kept identical to the main experiments.
\vspace{-0.5em}

\paragraph{Ablation variants.}
Since LIM's latent embeddings are in the \emph{same embedding space} as the language model inputs, we consider several simple substitutes:
(i) \textbf{Zero Embeddings}, where all visual-token slots are filled with the all-zero vector;
(ii) \textbf{Whitespace/Newline tokens}, where we repeat the embedding of a trivial, largely non-informative token (e.g., ``\texttt{ }'' or ``\texttt{\textbackslash n}'') across all visual slots;
and (iii) \textbf{Random tokens}, where we sample random vectors either (a) rescaled to match the text-embedding scale (per-dimension mean/variance or norm), or (b) directly from $\mathcal{N}(0,1)$; and \textbf{(iv) Blank image inputs}, where we provide a \emph{content-free} image (all-white or all-black) together with the same text prompt to the VLM, testing whether merely supplying an arbitrary visual input (without meaningful content) can yield similar gains.

\paragraph{Results.}
Across all variants, we find that \emph{arbitrary} visual tokens do \textbf{not} reproduce the improvements of LIM.
In most cases, these substitutes yield negligible gains over the text-only baseline, and in some settings they even \emph{degrade} both accuracy and calibration, suggesting that naively populating the missing-modality channel can introduce harmful spurious signals.
Figure~\ref{fig:ablation_randomness} summarizes ACC and ECE for each ablation, showing that LIM is the only approach that consistently improves both metrics.
This supports our central claim: LIM's benefits arise from \emph{structured, text-conditioned latent completion} rather than from the mere presence of additional tokens.

\section{Discussion}
\paragraph{Routing to an LLM vs.\ LIM.}
A natural alternative to our approach is to route text-only queries to a separate text-only LLM rather than augmenting the VLM with LIM.
While such routing could avoid the missing-modality mismatch, it introduces additional system complexity: it requires reliable modality detection, maintaining multiple models in deployment, and handling potentially inconsistent behaviors across models.

In contrast, our approach preserves a single unified VLM and augments it with a lightweight module, allowing the system to remain architecturally simple while improving calibration under missing modalities.
More importantly, our goal is not merely to recover text-only performance, but to investigate whether the multimodal latent space of a VLM can be leveraged even when one modality is absent.
LIM explicitly predicts a latent embedding compatible with the VLM's internal representation space, enabling the model to perform inference under missing-modality conditions without switching to a separate model.

\section{Conclusion}


We showed that vision-language models exhibit severe miscalibration when deployed under text-only inputs, even when accuracy remains relatively strong. This failure is not merely a consequence of semantic information loss, but reflects a systematic mismatch between multimodal training and text-only inference.
To address this issue, we proposed the Latent Imagination Module (LIM), which predicts task-relevant imagined representations from text and injects them into a frozen VLM backbone. LIM bridges the missing-modality gap and substantially improves both accuracy and confidence calibration in text-only deployment and missing-image settings, including on unseen benchmarks. Overall, our results highlight latent modality completion as a principled and efficient route to more reliable multimodal models under modality mismatch.

\newpage


\bibliography{iclr2026_conference}

\begin{thebibliography}{32}
\providecommand{\natexlab}[1]{#1}
\providecommand{\url}[1]{\texttt{#1}}
\expandafter\ifx\csname urlstyle\endcsname\relax
  \providecommand{\doi}[1]{doi: #1}\else
  \providecommand{\doi}{doi: \begingroup \urlstyle{rm}\Url}\fi

\bibitem[Antol et~al.(2015)Antol, Agrawal, Lu, Mitchell, Batra, Zitnick, and Parikh]{antol2015vqa}
Stanislaw Antol, Aishwarya Agrawal, Jiasen Lu, Margaret Mitchell, Dhruv Batra, C~Lawrence Zitnick, and Devi Parikh.
\newblock {VQA}: Visual question answering.
\newblock In \emph{Proceedings of the IEEE International Conference on Computer Vision (ICCV)}, pp.\  2425--2433, 2015.

\bibitem[Chatterji et~al.(2025)Chatterji, Cunningham, Deming, Hitzig, Ong, Shan, and Wadman]{chatterji2025people}
Aaron Chatterji, Thomas Cunningham, David~J Deming, Zoe Hitzig, Christopher Ong, Carl~Yan Shan, and Kevin Wadman.
\newblock How people use {ChatGPT}.
\newblock Technical report, National Bureau of Economic Research, 2025.

\bibitem[Chen et~al.(2025)Chen, Hu, He, Deng, Zhang, and Hong]{chen2025unveiling}
Zijun Chen, Wenbo Hu, Guande He, Zhijie Deng, Zheng Zhang, and Richang Hong.
\newblock Unveiling uncertainty: A deep dive into calibration and performance of multimodal large language models.
\newblock In \emph{Proceedings of the 31st International Conference on Computational Linguistics}, pp.\  3095--3109, 2025.

\bibitem[Clark et~al.(2018)Clark, Cowhey, Etzioni, Khot, Sabharwal, Schoenick, and Tafjord]{clark2018think}
Peter Clark, Isaac Cowhey, Oren Etzioni, Tushar Khot, Ashish Sabharwal, Carissa Schoenick, and Oyvind Tafjord.
\newblock Think you have solved question answering? try arc, the ai2 reasoning challenge.
\newblock arXiv preprint arXiv:1803.05457, 2018.

\bibitem[Deitke et~al.(2025)Deitke, Clark, Lee, Tripathi, Yang, Park, Salehi, Muennighoff, Lo, Soldaini, et~al.]{deitke2025molmo}
Matt Deitke, Christopher Clark, Sangho Lee, Rohun Tripathi, Yue Yang, Jae~Sung Park, Mohammadreza Salehi, Niklas Muennighoff, Kyle Lo, Luca Soldaini, et~al.
\newblock Molmo and pixmo: Open weights and open data for state-of-the-art vision-language models.
\newblock In \emph{Proceedings of the IEEE/CVF Conference on Computer Vision and Pattern Recognition (CVPR)}, pp.\  91--104, 2025.

\bibitem[Dolan \& Brockett(2005)Dolan and Brockett]{dolan2005automatically}
William~B Dolan and Chris Brockett.
\newblock Automatically constructing a corpus of sentential paraphrases.
\newblock In \emph{Proceedings of the Third International Workshop on Paraphrasing ({IWP}2005)}, 2005.

\bibitem[Guo et~al.(2017)Guo, Pleiss, Sun, and Weinberger]{guo2017calibration}
Chuan Guo, Geoff Pleiss, Yu~Sun, and Kilian~Q Weinberger.
\newblock On calibration of modern neural networks.
\newblock In \emph{Proceedings of the 34th International Conference on Machine Learning}, pp.\  1321--1330. PMLR, 2017.

\bibitem[Han et~al.(2022)Han, Chen, Kan, and Poria]{han2022mm}
Wei Han, Hui Chen, Min-Yen Kan, and Soujanya Poria.
\newblock Mm-align: Learning optimal transport-based alignment dynamics for fast and accurate inference on missing modality sequences.
\newblock arXiv preprint arXiv:2210.12798, 2022.

\bibitem[Hendrycks \& Gimpel(2016)Hendrycks and Gimpel]{hendrycks2016baseline}
Dan Hendrycks and Kevin Gimpel.
\newblock A baseline for detecting misclassified and out-of-distribution examples in neural networks.
\newblock arXiv preprint arXiv:1610.02136, 2016.

\bibitem[Hendrycks et~al.(2020)Hendrycks, Burns, Basart, Zou, Mazeika, Song, and Steinhardt]{hendrycks2020measuring}
Dan Hendrycks, Collin Burns, Steven Basart, Andy Zou, Mantas Mazeika, Dawn Song, and Jacob Steinhardt.
\newblock Measuring massive multitask language understanding.
\newblock arXiv preprint arXiv:2009.03300, 2020.

\bibitem[Heo et~al.(2018)Heo, Lee, Kim, Lee, Kim, Yang, and Hwang]{heo2018uncertainty}
Jay Heo, Hae~Beom Lee, Saehoon Kim, Juho Lee, Kwang~Joon Kim, Eunho Yang, and Sung~Ju Hwang.
\newblock Uncertainty-aware attention for reliable interpretation and prediction.
\newblock In \emph{Advances in Neural Information Processing Systems}, volume~31, 2018.

\bibitem[Hurst et~al.(2024)Hurst, Lerer, Goucher, Perelman, Ramesh, Clark, Ostrow, Welihinda, Hayes, Radford, et~al.]{hurst2024gpt}
Aaron Hurst, Adam Lerer, Adam~P Goucher, Adam Perelman, Aditya Ramesh, Aidan Clark, AJ~Ostrow, Akila Welihinda, Alan Hayes, Alec Radford, et~al.
\newblock {GPT}-4o system card.
\newblock arXiv preprint arXiv:2410.21276, 2024.

\bibitem[Jang et~al.(2025)Jang, Cho, Lee, Lee, and Lee]{jang2025reliable}
Chaeyun Jang, Deukhwan Cho, Seanie Lee, Hyungi Lee, and Juho Lee.
\newblock Reliable decision-making via calibration-oriented retrieval-augmented generation.
\newblock In \emph{The Thirty-ninth Annual Conference on Neural Information Processing Systems}, 2025.

\bibitem[Khot et~al.(2020)Khot, Clark, Guerquin, Jansen, and Sabharwal]{khot2020qasc}
Tushar Khot, Peter Clark, Michal Guerquin, Peter Jansen, and Ashish Sabharwal.
\newblock Qasc: A dataset for question answering via sentence composition.
\newblock In \emph{Proceedings of the AAAI Conference on Artificial Intelligence}, volume~34, pp.\  8082--8090, 2020.

\bibitem[Kuhn et~al.(2023)Kuhn, Gal, and Farquhar]{kuhn2023semantic}
Lorenz Kuhn, Yarin Gal, and Sebastian Farquhar.
\newblock Semantic uncertainty: Linguistic invariances for uncertainty estimation in natural language generation.
\newblock arXiv preprint arXiv:2302.09664, 2023.

\bibitem[Liu et~al.(2023)Liu, Li, Wu, and Lee]{liu2023visual}
Haotian Liu, Chunyuan Li, Qingyang Wu, and Yong~Jae Lee.
\newblock Visual instruction tuning.
\newblock In \emph{Thirty-seventh Conference on Neural Information Processing Systems}, volume~36, pp.\  34892--34916, 2023.

\bibitem[Liu et~al.(2020)Liu, Cui, Liu, Huang, Wang, and Zhang]{liu2020logiqa}
Jian Liu, Leyang Cui, Hanmeng Liu, Dandan Huang, Yile Wang, and Yue Zhang.
\newblock Logiqa: A challenge dataset for machine reading comprehension with logical reasoning.
\newblock arXiv preprint arXiv:2007.08124, 2020.

\bibitem[Lu et~al.(2022)Lu, Mishra, Xia, Qiu, Chang, Zhu, Tafjord, Clark, and Kalyan]{lu2022learn}
Pan Lu, Swaroop Mishra, Tanglin Xia, Liang Qiu, Kai-Wei Chang, Song-Chun Zhu, Oyvind Tafjord, Peter Clark, and Ashwin Kalyan.
\newblock Learn to explain: Multimodal reasoning via thought chains for science question answering.
\newblock In \emph{Advances in Neural Information Processing Systems}, volume~35, pp.\  2507--2521, 2022.

\bibitem[Ma et~al.(2023)Ma, Zhang, Zhang, Wu, Fu, Zhou, and Hu]{ma2023calibrating}
Huan Ma, Qingyang Zhang, Changqing Zhang, Bingzhe Wu, Huazhu Fu, Joey~Tianyi Zhou, and Qinghua Hu.
\newblock Calibrating multimodal learning.
\newblock In \emph{Proceedings of the 40th International Conference on Machine Learning}, pp.\  23429--23450. PMLR, 2023.

\bibitem[Naeini et~al.(2015)Naeini, Cooper, and Hauskrecht]{ece}
Mahdi~Pakdaman Naeini, Gregory Cooper, and Milos Hauskrecht.
\newblock Obtaining well calibrated probabilities using bayesian binning.
\newblock In \emph{Proceedings of the AAAI Conference on Artificial Intelligence}, volume~29, 2015.

\bibitem[Poudel et~al.(2025)Poudel, Chhetri, Gyawali, Leontidis, and Bhattarai]{poudel2025multimodal}
Pranav Poudel, Aavash Chhetri, Prashnna Gyawali, Georgios Leontidis, and Binod Bhattarai.
\newblock Multimodal federated learning with missing modalities through feature imputation network.
\newblock In \emph{Annual Conference on Medical Image Understanding and Analysis}, pp.\  289--299. Springer, 2025.

\bibitem[Schuster et~al.(2021)Schuster, Fisch, and Barzilay]{schuster2021get}
Tal Schuster, Adam Fisch, and Regina Barzilay.
\newblock Get your vitamin c! robust fact verification with contrastive evidence.
\newblock arXiv preprint arXiv:2103.08541, 2021.

\bibitem[Socher et~al.(2013)Socher, Perelygin, Wu, Chuang, Manning, Ng, and Potts]{socher2013recursive}
Richard Socher, Alex Perelygin, Jean Wu, Jason Chuang, Christopher~D Manning, Andrew~Y Ng, and Christopher Potts.
\newblock Recursive deep models for semantic compositionality over a sentiment treebank.
\newblock In \emph{Proceedings of the 2013 Conference on Empirical Methods in Natural Language Processing}, pp.\  1631--1642, 2013.

\bibitem[Sun et~al.(2024)Sun, Zhang, Han, Ruan, and Li]{sun2024redcore}
Jun Sun, Xinxin Zhang, Shoukang Han, Yu-Ping Ruan, and Taihao Li.
\newblock Redcore: Relative advantage aware cross-modal representation learning for missing modalities with imbalanced missing rates.
\newblock In \emph{Proceedings of the AAAI Conference on Artificial Intelligence}, volume~38, pp.\  15173--15182, 2024.

\bibitem[Talmor et~al.(2019)Talmor, Herzig, Lourie, and Berant]{talmor2019commonsenseqa}
Alon Talmor, Jonathan Herzig, Nicholas Lourie, and Jonathan Berant.
\newblock {CommonsenseQA}: A question answering challenge targeting commonsense knowledge.
\newblock In \emph{Proceedings of the 2019 Conference of the North {A}merican Chapter of the Association for Computational Linguistics: Human Language Technologies, Volume 1 (Long and Short Papers)}, pp.\  4149--4158, 2019.

\bibitem[Wang \& Zhou(2024)Wang and Zhou]{wang2024chain}
Xuezhi Wang and Denny Zhou.
\newblock Chain-of-thought reasoning without prompting.
\newblock In \emph{The Thirty-eighth Annual Conference on Neural Information Processing Systems}, volume~37, pp.\  66383--66409, 2024.

\bibitem[Warstadt et~al.(2019)Warstadt, Singh, and Bowman]{warstadt2019neural}
Alex Warstadt, Amanpreet Singh, and Samuel~R Bowman.
\newblock Neural network acceptability judgments.
\newblock \emph{Transactions of the Association for Computational Linguistics}, 7:\penalty0 625--641, 2019.

\bibitem[Xiong et~al.(2024)Xiong, Hu, Lu, LI, Fu, He, and Hooi]{xiongcan}
Miao Xiong, Zhiyuan Hu, Xinyang Lu, YIFEI LI, Jie Fu, Junxian He, and Bryan Hooi.
\newblock Can {LLMs} express their uncertainty? an empirical evaluation of confidence elicitation in llms.
\newblock In \emph{The Twelfth International Conference on Learning Representations}, 2024.

\bibitem[Xu et~al.(2025)Xu, Wang, Zhu, Pan, Chen, Fan, Chen, and Yu]{xu2025alignment}
Hongshen Xu, Zihan Wang, Zichen Zhu, Lei Pan, Xingyu Chen, Shuai Fan, Lu~Chen, and Kai Yu.
\newblock Alignment for efficient tool calling of large language models.
\newblock In \emph{Proceedings of the 2025 Conference on Empirical Methods in Natural Language Processing}, pp.\  17787--17803, 2025.

\bibitem[Yang et~al.(2024)Yang, Robeyns, Wang, and Aitchison]{yangbayesian}
Adam~X Yang, Maxime Robeyns, Xi~Wang, and Laurence Aitchison.
\newblock Bayesian low-rank adaptation for large language models.
\newblock In \emph{The Twelfth International Conference on Learning Representations}, 2024.

\bibitem[Yang et~al.(2025)Yang, Li, Yang, Zhang, Hui, Zheng, Yu, Gao, Huang, Lv, et~al.]{yang2025qwen3}
An~Yang, Anfeng Li, Baosong Yang, Beichen Zhang, Binyuan Hui, Bo~Zheng, Bowen Yu, Chang Gao, Chengen Huang, Chenxu Lv, et~al.
\newblock Qwen3 technical report.
\newblock arXiv preprint arXiv:2505.09388, 2025.

\bibitem[Zhang et~al.(2015)Zhang, Zhao, and LeCun]{zhang2015character}
Xiang Zhang, Junbo Zhao, and Yann LeCun.
\newblock Character-level convolutional networks for text classification.
\newblock \emph{Advances in Neural Information Processing Systems}, 28, 2015.

\end{thebibliography}
\bibliographystyle{iclr2026_conference}

\newpage
\appendix
\section{Related Works}
\label{related}


\paragraph{Uncertainty Estimation and Calibration in Modern Models}
\label{sec:llm calibration}
Model calibration has long been studied as a key component of reliable machine learning systems, revealing that high-capacity neural networks often exhibit systematic overconfidence despite strong predictive accuracy. Early work demonstrated that modern deep models are frequently miscalibrated and that simple post-hoc methods, such as temperature scaling, can significantly improve confidence accuracy alignment without sacrificing performance~\citep{guo2017calibration, heo2018uncertainty}. More recent studies extend these findings to LLMs, showing that model scaling and fine-tuning can further exacerbate miscalibration, particularly in generative settings. As a result, calibration techniques tailored to LLMs, including Bayesian uncertainty modeling and explicit confidence evaluation protocols, have been proposed to align predicted confidence more closely with the likelihood of true correctness~\citep{xiongcan, yangbayesian}. However, this line of work predominantly assumes text-only training and inference, leaving open questions regarding calibration behavior under more complex multimodal training regimes.


\paragraph{Uncertainty Calibration in Multimodal Models}
\label{sec:VLM calibration}
Parallel to advances in LLM calibration, recent work has explored uncertainty estimation and calibration in VLMs, typically focusing on settings where visual and textual inputs are jointly available at inference time~\citep{ma2023calibrating, chen2025unveiling}. These studies analyze confidence alignment in multimodal tasks such as visual question answering or captioning, implicitly assuming matched modality conditions between training and deployment. In practice, however, multimodal models are frequently applied to text-only tasks, even when their language backbone has been substantially adapted through multimodal training~\citep{chatterji2025people}. Such adaptation can fundamentally alter internal representations and uncertainty estimation mechanisms, rendering text-only inputs out-of-distribution with respect to the multimodal training regime. As a consequence, calibration on text-only tasks can degrade severely, even compared to the original text-only backbone, an issue that remains largely unexplored in prior calibration literature. This overlooked modality mismatch motivates methods that explicitly address calibration degradation caused by missing or imbalanced modalities.

\paragraph{Missing-modality robustness in multimodal learning}
Prior work on missing modalities in multimodal learning has largely followed two strategies.
First, \emph{robust representation and alignment} methods train models to remain stable when modalities are intermittently absent, e.g., by aligning cross-modal dynamics or using denoising objectives so that inference degrades gracefully under missing inputs~\citep{han2022mm}.
Second, \emph{feature/latent-level imputation} methods explicitly predict the missing-modality in an intermediate representation space (rather than generating pixels), and use the reconstructed features to recover downstream performance under missingness~\citep{sun2024redcore, poudel2025multimodal}.

Our setting differs in that we study \emph{text-only deployment of unified VLMs} and target \emph{confidence calibration} in addition to accuracy.
Accordingly, LIM performs text-conditioned \emph{latent completion} at the VLM interface: it directly predicts the latent embedding block expected by the backbone and injects it during inference, avoiding expensive pixel-level generation while explicitly correcting missing-modality over-confidence.

\label{appendix}
\section{Additional Details on Empirical Findings}

\subsection{Implementation Details for Image-to-Text Description Generation}
\label{app:description generation}

To construct the \textsc{Description+Question} condition in Section~\ref{sec:finding_caption_and_regen}, we generate a high-fidelity textual description for each VQA image using an off-the-shelf vision-language model.
Specifically, we use \textbf{LLaVA-1.5-7B} (\texttt{llava-hf/llava-1.5-7b-hf}) as the captioning model, and prompt it to produce a detailed, attribute-rich description.

\paragraph{Prompt.}
We use the following instruction prompt, designed to elicit fine-grained visual attributes that may be relevant for downstream question answering:

\begin{center}
\setlength{\fboxsep}{8pt}    
\setlength{\fboxrule}{0.6pt} 
\fbox{%
\begin{minipage}{0.95\linewidth}
\small
\textbf{Prompt:} Describe this image in detail. Include information about the setting, objects, people, clothing, actions, facial expressions, colors, numbers, and any notable background elements. Be as specific and thorough as possible.
\end{minipage}}
\end{center}

\paragraph{Generation settings.}
Descriptions are generated with greedy decoding (\texttt{do\_sample=False}) and a maximum length of \texttt{200} new tokens (\texttt{max\_new\_tokens=200}).
All generations are performed in \texttt{float16} with automatic device placement (\texttt{device\_map="auto"}). We decode model outputs with \texttt{skip\_special\_tokens=True}.

\paragraph{Example.}
Figure~\ref{fig:caption_example} shows an example image and the corresponding generated description.

\begin{figure}[t]
    \centering
    \setlength{\fboxsep}{8pt}
    \setlength{\fboxrule}{0.6pt}

    \begin{minipage}[c]{0.48\linewidth} 
        \centering
        \includegraphics[width=\linewidth]{figs/description_ex.pdf}
    \end{minipage}
    \hfill
    \begin{minipage}[c]{0.48\linewidth} 
        \small
        \fbox{%
        \begin{minipage}{0.96\linewidth}
        {\textbf{Text Description for the image}}\\[4pt]
        The image depicts a lively outdoor market scene with several people shopping for fresh produce. There are at least nine people visible in the scene, with some standing near the produce stands and others browsing the market.
        The market is filled with a variety of fruits and vegetables, including multiple apples, oranges, and carrots. The apples are scattered throughout the market, with some near the center and others closer to the edges. Oranges can be found in different areas of the market, with some near the center and others closer to the edges. Carrots are also abundant, with several bunches placed in various locations throughout the market.
        The market appears to be a popular destination for shoppers, as evidenced by the large number of people present and the abundance of fresh produce available.
        \end{minipage}}
    \end{minipage}

    \caption{\textbf{Example of description generation.} \textbf{(Left)} Input image. \textbf{(Right)} LLaVA-generated description produced by the prompt in Appendix~\ref{app:description generation}.}
    \label{fig:caption_example}
\end{figure}

\subsection{Additional Model Family: Qwen-Based VLMs under Text-Only Inputs}
\label{app:qwen_additional}

In Section~\ref{sec:finding_backbone_gap}, we show that LLaVA does not reliably fall back to its language backbone under text-only prompting.
To verify that this phenomenon is not specific to the Vicuna/LLaVA family, we additionally examine a Qwen-based model family under the same \emph{text-only} prompting setup.

\paragraph{Models and setup.}
We compare a text-native Qwen LLM, \textbf{Qwen2.5-7B}, against its Qwen-based VLM counterpart,
\textbf{Qwen2.5-VL-7B-Instruct},
using identical text inputs (question, options, and instructions) and matched decoding settings.
For consistency with the main analysis, we use our primary logit-based confidence estimator (MSP) and report both accuracy and calibration error (ECE).

\begin{figure}[t]
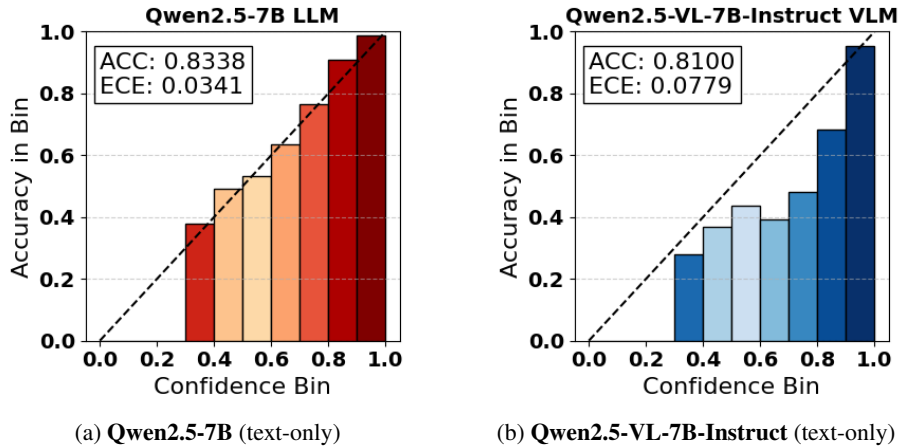

    \centering
    \begin{minipage}{0.40\columnwidth}
        \centering
        \includegraphics[width=\linewidth]{figs/qwen2.5_7b_llm.pdf}
        \small (a) \textbf{Qwen2.5-7B} (text-only)
    \end{minipage}
    \hspace{0.05\columnwidth}
    \begin{minipage}{0.40\columnwidth}
        \centering
        \includegraphics[width=\linewidth]{figs/qwen2.5_vl_7b_vlm.pdf}
        \small (b) \textbf{Qwen2.5-VL-7B-Instruct} (text-only)
    \end{minipage}
    \caption{\textbf{Reliability diagrams for Qwen-based models under text-only prompting.}
    We compare \textbf{Qwen2.5-7B} and \textbf{Qwen2.5-VL-7B-Instruct} using MSP confidence.
    The Qwen-based VLM shows a similar over-confidence trend under missing vision modality as observed for LLaVA in Section~\ref{sec:finding_backbone_gap}.}
    \label{fig:qwen_text_only_rel}
\end{figure}

\paragraph{Results.}
Table~\ref{tab:qwen_text_only_calibration} reports the aggregated ACC/ECE for \textbf{Qwen2.5-7B} and \textbf{Qwen2.5-VL-7B-Instruct} under text-only prompting.
Consistent with our main findings, the Qwen-based VLM exhibits worse calibration than the corresponding language backbone when the visual modality is absent.
Figure~\ref{fig:qwen_text_only_rel} further visualizes this behavior via reliability diagrams.

\begin{table}[!t]
\centering
\caption{\textbf{Qwen-based family under text-only prompting (MSP confidence).}
Comparison between \textbf{Qwen2.5-7B} and \textbf{Qwen2.5-VL-7B-Instruct}.
}
\vspace{1mm}
\renewcommand{\arraystretch}{1.0}
\setlength{\tabcolsep}{7pt}
\begin{tabular}{l|cc}
\toprule
\textbf{Model} & \textbf{ACC} $\uparrow$ & \textbf{ECE} $\downarrow$ \\
\midrule
\textbf{Qwen2.5-7B} (\texttt{Qwen/Qwen2.5-7B}) & 83.38 & 0.0341 \\
\textbf{Qwen2.5-VL-7B-Instruct} (\texttt{Qwen/Qwen2.5-VL-7B-Instruct}) & 81.00 & 0.0779 \\
\bottomrule
\end{tabular}
\label{tab:qwen_text_only_calibration}
\end{table}

\subsection{Prompting Details for Diffusion-Generated ``Helpful Images''}
\label{app:helpful_image_prompt}

In Section~\ref{sec:finding_text_task_with_generated_images}, we generate an auxiliary image for each text-only QA instance and feed it to the VLM as a visual input.
This appendix details the image-generation prompt template and diffusion hyperparameters.

\paragraph{Text-to-image model.}
We use Stable Diffusion v2.1 base (\texttt{stabilityai/stable-diffusion-2-1-base}) via the \texttt{diffusers} \texttt{StableDiffusionPipeline}.
All generations are performed in \texttt{float16} on GPU.
We enable attention slicing (\texttt{enable\_attention\_slicing}) to reduce VRAM usage.

\paragraph{Prompt template for helpful images.}
For each QA instance, we build a diffusion prompt from the question text.
Let $q$ denote the instance question string.
We use the following fixed template:
\begin{center}
\setlength{\fboxsep}{8pt}
\setlength{\fboxrule}{0.6pt}
\fbox{%
\begin{minipage}{0.95\linewidth}
\small
\textbf{Helpful-image prompt template:}\\
\texttt{Given the following question [q], generate an abstract yet informative image that expresses the essential concepts and mathematical reasoning required to solve the problem. Focus on intuitive shapes, spatial relationships, and symbolic representations that make the solution approach clear.}
\end{minipage}}
\end{center}
This template encourages abstract, concept-focused visualizations (e.g., diagrams, symbolic shapes, spatial relations) rather than literal scene rendering.

\paragraph{Diffusion hyperparameters.}
We generate one image per instance using \texttt{num\_inference\_steps=40} (unless otherwise noted).
Other sampling parameters (e.g., guidance scale, image resolution, random seed) are left as pipeline defaults.

\end{document}